A unified spectra analysis workflow for the assessment of microbial contamination of ready-to-eat green salads: Comparative study and application of non-invasive sensors

Panagiotis TSAKANIKAS*, Lemonia-Christina FENGOU, Evanthia MANTHOU, Alexandra LIANOU, Efstathios Z. PANAGOU & George – John  E. NYCHAS*

*Laboratory of Microbiology and Biotechnology of Foods, Department of Food Science and Human Nutrition, School of Food, Biotechnology and Development, Agricultural University of Athens, 11855, Athens, Greece*

Panagiotis Tsakanikas*: p.tsakanikas@aua.gr

Lemonia-Christina Fengou: lefengou@gmail.com

Evanthia Manthou: evita.m@windowslive.com

Alexandra Lianou: alianou@aua.gr

Efstathios Z. Panagou: stathispanagou@aua.gr

George-John E. Nychas*: gjn@aua.gr

*\* Authors for correspondence*

**Abstract**

The present study provides a comparative assessment of non-invasive sensors as means of estimating the microbial contamination and "time-on-shelf" (i.e. storage time) of leafy green vegetables, using a novel unified spectra analysis workflow. Two types of fresh ready-to-eat green salads were used in the context of this study for the purpose of evaluating the efficiency and practical application of the presented workflow: rocket and baby spinach salads. The employed analysis workflow consisted of robust data normalization, powerful feature selection based on random forests regression, and selection of the number of partial least squares regression coefficients in the training process by estimating the "knee-point" on the explained variance plot. Training processes were based on microbiological and spectral data derived during storage of green salad samples at isothermal conditions (4, 8 and 12°C), whereas testing was performed on data during storage under dynamic temperature conditions (simulating real-life temperature fluctuations in the food supply chain). Since an increasing interest in the use of non-invasive sensors in food quality assessment has been made evident in recent years, the unified spectra analysis workflow described herein, by being based on the creation/usage of limited sized featured sets, could be very useful in food-specific low-cost sensor development.

1. **Introduction**

In recent years, the production, sale and consumption of fresh raw fruits and vegetables, especially as pre-cut and ready-to-eat (minimally processed by washing, slicing or shredding and packaging) salads have undergone substantial increases in the European Union as well as the United States. This is the result of consumer preferences for fresher, more convenient and nutritious foods that meet the needs of busier lifestyles, at least in the developed countries. Vegetables are highly contaminated by nature due either to agronomic systems employed in their production (e.g., irrigation with contaminated water, organic fertilizers such as manure, etc.) or during processing, handling and marketing, (Francis et al., 2012; Skandamis & Nychas, 2011). Nonetheless, consumers demand fresh produce commodities, which should be not only perfectly safe for human consumption, but visually attractive as well.

Over the past decade, a number of sensors based on vibrational spectroscopy or hyperspectral/multispectral imaging have been developed (Rateni, Dario, & Cavallo, 2017). Although such sensors have started gaining popularity as rapid and efficient methods for assessing food quality and/or food composition, their utilization in handling of microbiological issues related to the current EU legislation (Commission, 2005) still remains to be firmly established. Indeed, the challenge of using non-invasive sensors as sensible alternatives to the costly and time-consuming conventional microbiological techniques has not been adequately tackled yet.

Spectroscopy which indeed allows rapid, non-invasive and non-destructive analysis (Nychas, Panagou, & Mohareb, 2016), cannot be considered as target-specific method and it must be associated with complex data analytics to extract the relevant microbiological and biochemical information. This can be achieved by the selections of the wavelength range for source and detector, and of the measurement setup. It is evident that due to the multi-dimensional nature of the data generated from such analyses, the output needs to be coupled with a suitable statistical approach or machine-learning algorithms before the results can be interpreted (Ropodi, Panagou, & Nychas, 2016). Choosing the optimum pattern recognition or machine learning approach for a given analytical platform is often challenging and involves a comparative analysis of various algorithms in order to achieve the best possible prediction accuracy (Estelles-Lopez et al., 2017; Tsakanikas, Pavlidis, & Nychas, 2015; Tsakanikas, Pavlidis, Panagou, & Nychas, 2016).

Here in, we introduce a unified data analysis and prediction model building workflow for ready-to-eat green salads (specifically rocket and baby spinach). To the best of our knowledge this is the first attempt for a unified data analysis approach applied to a whole category of food. This unified approach by itself consists a major contribution to the field of microbial contamination assessment on food. In addition, coupled with the non-invasive sensors used (a nowadays necessity for food quality assessment and monitoring), namely FTIR, NIR and VIS, we further enhance its impact and applicability. Further aspects of possible contributions via the developed feature selection process (selection of specific wavelengths of the sensors) may lead to the development of sensors focus and targeted to specific food types offering optimized precision and maybe lower cost for the industry.

In the present study for the first time, to our knowledge, three sensors of which two are commercially available and one developed within the frame of PhasmaFOOD EU project (http://www.phasmafood.eu/ further details are provided in the Materials & Methods section) were used for the assessment of microbiological quality and freshness (storage time) of fresh produce commodities. The acquired measurements for the three abovementioned sensors were analyzed using advanced data analysis to predict bacterial counts of selected vegetable products (i.e. rocket and baby spinach salads), with the latter being stored under aerobic isothermal as well as dynamic temperature conditions. Towards a meaningful predictive model, although trained and build at isothermal storage conditions dataset, the testing has been performed on samples stored at storage temperatures with dynamic variations, and thus simulating real life conditions.

**2. Materials & Methods**

2.1. Green salad samples and storage conditions

Fresh and ready-to-eat rocket and baby spinach salads (i.e. sealed plastic bags each one containing 85 g or 125 g of fresh-cut, washed and drained rocket or baby spinach leaves, respectively) were obtained from a local manufacturer and transported to the laboratory within 24 h from production. The salads were stored in their original packages (i.e. aerobic storage) at three different isothermal conditions (4, 8 and 12°C), as well as at dynamic storage conditions with periodic temperature changes from 4 to 12°C (8 h at 4°C, 8 h at 8°C and 8 h at 12°C). Sample storage took place in high-precision (±0.5°C) programmable incubators (MIR-153, Sanyo Electric Co., Osaka, Japan) for a maximum time period of approximately 11 days (i.e. 254 h), and the incubation temperatures were recorded at 15-min intervals throughout storage using electronic temperature-monitoring devices (COX TRACER®, Cox Technologies Inc., Belmont, NC, USA). On the day of arrival to the laboratory (time-zero) and at regular time intervals during storage, depending on the applied storage temperature, duplicate salad samples (originating from different packages) were subjected to the following analytical procedures: (i) microbiological analyses and pH measurements; (ii) Fourier transform infrared (FTIR) spectroscopy measurements; (iii) near infrared (NIR) spectroscopy measurements; and (iv) visible (VIS) spectroscopy measurements. Two independent experiments (i.e. different time instances and different salad batches) were conducted, and a total of 217 rocket salad and 212 baby spinach salad samples were subjected to the aforementioned analyses during storage at the different temperature conditions.

2.2. Microbiological analyses

A 25-g portion of the content of a salad package was transferred aseptically to a 400-ml sterile stomacher bag (Seward Medical, London, UK) containing 225 ml of sterilized quarter-strength Ringer's solution (Lab M Limited, Lancashire, UK), and homogenized in a Stomacher apparatus (Lab Blender 400, Seward Medical) for 60 sec at room temperature. For the enumeration of the total mesophilic microbial populations (total viable counts, TVC), appropriate serial decimal dilutions in Ringer's solution were surface plated on tryptic glucose yeast agar (Plate Count Agar; Biolife, Milan, Italy), and colonies were counted after incubation of plates at 25°C for 72 h. The obtained microbiological data were converted to log (colony forming units) per gram of vegetable salad (log CFU/g). Upon completion of the microbiological analyses, the pH values of the samples also were measured using a digital pH meter (RL150, Russell pH, Cork, Ireland) with a glass electrode (Metrohm AG, Herisau, Switzerland).

2.3. Sensors

2.3.1. FTIR spectroscopy

FTIR spectral data were collected using a ZnSe 45° HATR (Horizontal Attenuated Total Reflectance) crystal (PIKE Technologies, Madison, Wisconsin, USA), and an FTIR-6200 JASCO spectrometer (Jasco Corp., Tokyo, Japan) equipped with a standard sample chamber, a triglycine sulphate (TGS) detector and a Ge/KBr beamsplitter. A small portion from each salad sample was cut in small pieces and transferred to the crystal plate, covered with a small piece of aluminum foil, and then pressed with a gripper to ensure the best possible contact with the crystal. The crystal used has a refractive index of 2.4 and a depth of penetration of 2.0 µm at 1000 cm$^{-1}$. Using the Spectra Manager™ Code of Federal Regulations (CFR) software version 2 (Jasco Corp.), spectra were collected over the wavenumber range of 4000 to 400 cm$^{-1}$, by accumulating 100 scans with a resolution of 4 cm$^{-1}$ and a total integration time of 2 min. Prior to the measurements of the tested samples, reference spectra were acquired using the cleaned blank (no added salad sample) crystal. After each measurement, the crystal's surface was cleaned, first with detergent and distilled water and then with analytical grade acetone, and dried using lint-free tissue. The FTIR spectra that were ultimately used in further analyses were in the approximate wavenumber range of 2700 to 900 cm$^{-1}$.

2.3.2. NIR spectroscopy

Spectra from the green salad samples were acquired using a NIR spectrometer (SGS1900), developed by Fraunhofer IMPS (Institute für Photonische Mikrosysteme, Dresden, Germany). The NIR spectrometer covers a wavelength range from 1000 to 1900 nm, and it is comprised of a MEMS-based scanning grating for spectral dispersion and an uncooled InGaAs diode for detection (Pügner, Knobbe, Grüger, & Schenk, 2012). A halogen light source providing illumination (in-house constructed by IPMS) was coupled via SMA 905 connector into an Ocean optics Y-shaped fiber bundle (QR400-ANGLE-VIS). The fiber bundle comprises of a probe tip with a window, which is set at 30° to the front face of the fiber. During measurement, this window is placed in direct contact with the sample. Light from the halogen source is fed to the sample and the diffuse reflectance from the sample is collected back into a 400 µm core optical fiber in the center of the fiber bundle. This fiber then transmits the collected light to the SGS1900 spectrometer via SMA connector. The software "Quickstep" (Hiperscan GmbH) served as operation software for the NIR spectroscopy measurements. For each salad sample, ten (10) spectral measurements (absorbance values) were acquired by placing the fiber's probe tip on different spots of the vegetable leaves' surface, and then their average value was used in order to combat and embed the inherent large diversity of the samples' absorbance values.

2.3.3. VIS spectroscopy

The VIS spectrometer sensor used in this study was the Hamamatsu C12880MA (Hamamatsu Photonics K.K., Shizuoka, Japan). The device has a spectral range from 340 to 850 nm, high sensitivity and spectral resolution of 15 nm. It should be noted that only visible range spectroscopy has been used, and in order for this to be ensured, a UV filter with a cutoff at 400 nm was introduced in front of the spectrometer aperture to avoid any signal interference. The salad samples were placed in Petri dishes, and, similarly to the NIR measurements, ten (10) different spectral measurements (absorbance values) were acquired at different spots of the samples and their average value was used in order for the inherent absorbance diversity to be taken into account.

2.4. Data analysis methodology

This section provides a detailed description of the data analysis workflow, which is, more or less, common for all studied sensors. The processing pipeline consists of feature selection (specific wavelengths/wavenumbers) on the basis of random forests (RFs) regression ensemble (Breiman, 2001) followed by partial least squares (PLS) regression coupled with automated selection of number of components.

Prior to feature selection, spectra were normalized under the Standard Normal Variate (SNV) normalization scheme (Barnes, Dhanoa, & Lister, 1989) and more specifically its robust version RNV (Guo, Wu, & Massart, 1999) (Eq. 1). In addition to enhancing data quality, reducing the correlated information across the different wavelengths/wavenumbers and eliminating the inherent (due to the acquisition process) multiplicative noise, the robust version of SNV also gives more reasonable results (i.e. without artifacts) and leads to improved analysis:

$$s_i^{snv} = \frac{s_i - \text{median}(S)}{mad(S)} \quad (1)$$

where *S* is the ensemble of all spectra, and $s_i$ and $s_i^{snv}$ the *i*th and the corresponding normalized spectra, respectively. Median Absolute Deviation (*mad*) (Hoaglin, Frederick Mosteller, & Tukey, 2000), is a robust measure of the variability of a univariate sample of quantitative data $s_1, s_2, ..., s_n$ computed as:

$$mad = median(|s_i - median(S)|) \quad (2)$$

After the spectra normalization and prior to regression of microbial contamination (TVC) and storage time, a feature selection step was introduced. This is considered critical not only in this specific case but in every regression/classification problems where small amounts of samples in high dimensional space (i.e. large number of variables) are available. So, in order to surpass this issue and not fall into overfitting, variable/feature set has to be decreased in a way that meaningful features are preserved while irrelevant (to the prediction of microbial contamination and storage time) and redundant ones are excluded. Several algorithms exist for feature selection, with most of them lying into the concept of feature ranking (James, Witten, Hastie, & Tibshirani, 2013) and selecting a user-defined number of features. Nevertheless, no adequate prior knowledge from the current scientific literature exists on wavelengths/wavenumbers indicative for microbial contamination of vegetables. Hence, apart from the fact that feature ranking algorithms do not perform well in terms of computational time and that the resulting feature subsets are redundant, it is also difficult to set a specific, even intuitive, threshold for the number of features to keep.

In this context, RFs were selected to be employed for regression (Breiman, 2001), which are an ensemble learning method for regression that constructs a multitude of decision trees at training time while provides as output the mean prediction (regression) of the individual trees. One justification for this selection is that RFs correct decision trees' habit of overfitting to their training set, something which is needed here as stated earlier. In addition, RFs under boosting (Duffy & Helmbold, 2002) scheme was employed for learning which has the advantage over averaging (Breiman, 2001) or bagging (Tin Kam, 1998), since boosting algorithms consist of iteratively learning weak classifiers with respect to a distribution and adding them to a final strong classifier. Then, the data are reweighted: examples that are misclassified gain weight and examples that are classified correctly lose. Thus, future weak learners focus more on the examples that previous weak learners misclassified, thus boosting their performance. Having in mind the aforementioned advantages of RFs, their application on the selected data is expected to output concrete and representative (yet not redundant) features from each sensor that best represents the samples in terms of discriminating the inherent microbial burden. Herein, the regression tree ensemble was trained using LSBoost (gradient boosting strategy applied for least squares) (Friedman, 2001) and 100 learning cycles, with all constant temperatures (i.e. 4, 8 and 12 $^0$C) being used as training set and the set of samples stored at dynamic temperature conditions being utilized as external test set. To justify and support this data partitioning for training and testing: our aim is to develop temperature-agnostic prediction models that perform well in real life applications, meaning dynamically changing temperatures of storage.

Finally, after the selection of features (i.e. spectra wavelengths/wavenumbers) and using only the resulted reduced in dimensions datasets, PLS regression (Wold, Sjöström, & Eriksson, 2001) was performed, since this method is a popular method in food quality applications (Panagou, Papadopoulou, Carstensen, & Nychas, 2014; Papadopoulou, Panagou, Tassou, & Nychas, 2011). A significant issue concerning PLS modeling, and specifically the optimum number of components selection, is the variation of the irrelevant to the phenomenon under study data, causing the inclusion of components that may have negative impact on the model. The selection of the optimum number of PLS components is made using the plot of variance to number of model coefficients at the "knee-point". The "knee-point" is defined as the intersection of two lines that are obtained as a result of an iterative curve fitting procedure. Specifically, by "walking" along the curve (a fitted smoothing spline (Chapra & Canale, 2006) to the posteriors data) from left to right one bisection point at a time and fitting two lines, one to all points to the left of the bisection point and another to all points to the right of the bisection point, the knee is the bisection point for which the sum of the Root Mean Square Errors (RMSE) for the two lines is minimized. The training process is performed on the dataset of samples stored at constant temperatures (for each produce commodity type independently), while 10-fold cross validation and 10 Monte-Carlo repartitions of the data were performed for model calibration. At this point we should state that the aforementioned procedure was applied not only for microbial contamination but also for "time-on-shelf" estimation and prediction. The term "time-on-shelf", is used to denote the time of the products' disposal after their production and which, in the context of the present study, is approximated by the storage time of the green salads.

## 3. Results & Discussion

As described in section 2.4, the training of the proposed data analysis workflow was performed on the samples stored at constant temperatures, i.e. 4, 8 and 12 $^0$C, while the test of the resulting prediction model on the data corresponding to the dynamic temperature storage conditions. Figure 1 shows the box plots corresponding to the spread of the TVC values for rocket (1A) and spinach (1B) respectively, throughout the storage experiment.

3.1. Dimensionality reduction – feature selection

The results of the methodology outlined earlier for all the studied sensors towards a unified feature selection scheme for spectral data analysis via the current state-of-art method for food quality assessment as the final step, i.e. PLS regression, are presented in this section. Following the RFs-based procedure for feature selection, as delineated in section 2.4, we get feature sets that are very limited compared to the original number of variables (Table 1). The reduced data dimensionality is to the profit of model robustness since the number of features are less/comparable to the number of samples, overfitting is avoided and also features that do not correlate to the microbial population increment along storage time are excluded from analysis, eliminating the bias of other than microbiological factors into the model. Thus, the resulting model is less sensitive to sample diversity. On the other hand, this increased generalization property of the model is reflected to the regression accuracy (as discussed next) mainly due the non-overfitting to the data.

3.2. Rocket salad case study

Concerning the rocket case study conducted in the framework of this research, the linear regression in terms of $y = ax + b$ first order polynomial fitting between the predicted and the actual TVC values determined by classical microbiological analysis is presented in Table 2 and Figure 2. This type of regression is used in order to evaluate and quantify the performance of the prediction models. As mentioned earlier, the developed PLSR model for each sensor (and case study) takes as inputs, i.e. predictors, the limited/decreased features resulting from RFs application during the training process. Figure 2 (A1, B1, C1) displays the regression line for each sensor spectral data where the dashed lines exhibit the 95% prediction boundaries. The slope parameter $a$ for all sensors shows a good correlation/correspondence between predicted and actually measured TVC values (ranging from 0.73 to 0.82) which is also suggested by the goodness-of-fit measure of R-square. These results suggest that the phenomenon under study, i.e. microbial contamination of rocket salad, can be assessed by sensors at the spectral range from visible to infrared, while information fusion across those wavelengths may result to improvement of model accuracy and robustness. It should also be mentioned that the RMSE values for every sensor is below 1 log CFU/g, which is a good result and acceptable for application in food microbiology, since deviations in TVC of approximately 0.5 log cycles is rather common even within the same laboratory.

In addition, analysis concerning the TVC values without the RFs feature selection step, but keeping the other processing steps introduced in the developed analysis workflow, also was performed. In the context of this analysis, data normalization with robust SNV and selection of the number of PLS components via the "knee-point" estimation were performed, as described in section 2.4. The results of this analysis clearly demonstrated the performance boost of the models developed after efficient features extraction (Table 2).

Apart from the analysis in terms of TVC prediction, we also performed an analysis in order to predict the "time-on-shelf", which is approximated by the storage time as defined earlier. Again, the same pipeline of analysis has been applied and followed, only this time the response values were the storage time sampling (i.e. measurement) intervals (please refer to Table 2 and Figure 2 (A2, B2, C2)). Here, the main result was that the FTIR sensor exhibited the best prediction with a deviation of approximately 17 h, while the slope $a$ and the R-square value were rather close to the ideal 1-to-1 relationship between actual and predicted storage time.

3.3. Baby spinach salad case study

In the case of the baby spinach case study, the results from the aforementioned analysis are summarized and presented in Table 3 and Figure 3, showing again the linear regression in terms of $y = ax + b$ between the predicted and the measured TVC values by classical microbiological analysis. Figure 3 (A1, B1, C1) displays the regression line for each sensor's spectral data where the dashed lines exhibit the 95% prediction boundaries. A major issue in the baby spinach samples, not encountered in the rocket salad case, was the considerably narrow TVC range (Figure 1B). Indeed, the spinach samples (even the ones corresponding to the beginning of the product's commercial shelf life and to the time-zero of microbiological analysis) had a high microbial concentration, frequently exceeding 7 logCFU/g, with the vast majority of values lying in a range of less than 1 log CFU/g. The high prevalence of microbial contamination in fresh spinach has been frequently reported in the scientific literature, not only in the case of raw produce but also in minimally-processed ready-to-eat salads (Mritunjay & Kumar, 2017; Oliveira, Maciel de Souza, Morato Bergamini, & De Martinis, 2011). However, as anyone can argue the acquired narrow-ranged dataset is not ideal for regression purposes or even for any other type of analysis since an estimation of a fixed TVC value for every new sample at *ca.* 8 log CFU/g would be a "safe" and not far from reality guess. Nevertheless, herein the data were analyzed as described previously. The outcomes concerning the slope parameter *a* for FTIR indicated a good correlation/correspondence between predicted and actually measured TVC (~0.74), while in the case of the NIR and VIS sensors the slope was inferior. It should be also mentioned that the RMSE values for all the studied sensors were below 0.5 log CFU/g which is a rather acceptable result for food microbiology applications as discussed previously. These results, although poor in terms of prediction to measured (actual) values apart from the FTIR sensor, normally would not suggest an acceptable predictive model for microbial contamination of baby spinach salad. Taking into account the limited range of TVC values where the regression line is very difficult to fit well on the data, and the fact that the RMSE values were less than the deviation of microbiological measurements within the laboratory measurements, we can argue that the models are at least better than an estimation of 0.8 log CFU/g for all samples, as stated earlier. So, it can be acclaimed that in the strict context of microbiology, the developed data analysis methodology not only is optimum (for such a limited TVC values' range) but also is applicable as a predictive model. To further support such arguments, the residuals of the predicted minus measured TVC values were also analysed (Fig. 4), and a distribution with mean value in the range (-0.01, 0.01) and a standard deviation in the approximate range (0.3, 0.5) was exhibited. Again, as when RMSE is considered, if the residuals' standard deviation is taken into account, it can be safely supported that the developed models can be used in practice, since the deviation within the laboratory measurements for TVC values is expected to be ±0.5 log cycles. Again, as in the previous subsection, the regression results of the model developed with the intermediate step of features selection (but with the rest processing steps introduced herein being maintained) is presented (Table 3).

Similarly to what was the case for the rocket salad case study, the same analysis was applied in order to predict the "time-on-shelf" as defined earlier. The results clearly support the justification and the argument made concerning the TVC prediction results. Here the range of responses for the regression is not limited as in the TVC case, and the correlation between predicted "time-on-shelf" and the actual storage time sampling interval is almost perfect. This can be assessed by both the slope parameter, which was >0.7 for all studied sensors, and the goodness-of-fit as this is demonstrated by the R-square values (please refer to Table 3 and Figure 3 (A2, B2, C2)).

## 4. Conclusions

The significance of an efficient variable decrement is of crucial importance for model training, especially when sample size is much smaller but it is also very significant for several other purposes. The reduced set of features, if efficiently extracted, is considered as de-correlated not only among the features but also to the phenomenon under study. Also the inherent redundant information is minimized. All the above result to more robust and accurate regression model development than using the full list of features. In addition, the creation/usage of limited sized featured sets could be very useful in food-specific low-cost sensor development since limited range of wavelengths would be required.

Apart from reduction of the number of the wavelengths needed for a robust and accurate prediction of microbial contamination and "time-on-shelf" estimation, another major outcome of the presented study is that not only we used different batches of samples (i.e. biological variability is taken into account), the testing of the models was performed on samples that were stored in dynamic temperature conditions, a simulation of what is really happening to ready-to-eat salads during their disposal in the food supply chain. So, it is justifiable to argue that the proposed workflow is applicable to real life, after succeeding in "difficult" data (simulating real life), and is expected to provide fruitful insights, via the in depth analysis of the wavelengths/wavenumbers selected into food microbial contamination evolution during storage.

Future research goals include in depth analysis of the selected wavelengths in order to correlate them with metabolites associated with the spoilage of fresh green salads, as well as exploration of data fusion strategies integrating the studied sensors. Finally, the potential application of the developed spectra analysis workflow to other types of food (e.g., animal origin foods) could also be assessed in the context of future research.

**Acknowledgements**

This work has been supported by the project "PhasmaFOOD", funded from the European Union's Horizon 2020 research and innovation programme under grant agreement No 732541.

| Sensor | Wavelength/wavenumber | Resolution | # variables | # selected features |
|---|---|---|---|---|
| **FTIR** | [1000-3000] cm$^{-1}$ | 4 cm$^{-1}$ | 2075 | 87 (rocket) 95 (spinach) |
| **NIR** | [1000-1900] nm | 1 nm | 901 | 94 (rocket) 91 (spinach) |
| **VIS** | [340-850] nm | 15 nm | 288 | 78 (rocket) 74 (spinach) |

**Table 1**. Summary of the sensors' properties, total number of variables and number of variables resulting after RFs-based feature selection.

| Microbial Contamination | a | B | R-square | RMSE |
|---|---|---|---|---|
| FTIR | 0.822 | 1.215 | 0.511 | 0.758 |
| NIR | 0.792 | 1.564 | 0.945 | 0.315 |
| VIS | 0.730 | 1.530 | 0.946 | 0.330 |
| **Microbial Contamination - No Feature Selection Performed** | | | | |
| FTIR | 0.633 | 2.485 | 0.416 | 0.687 |
| NIR | 0.570 | 3.067 | 0.118 | 0.865 |
| VIS | 0.420 | 4.086 | 0.944 | 0.249 |
| | | | | |
| **Time-on-shelf** | a | b | R-square | RMSE |
| FTIR | 0.984 | 14.490 | 0.958 | 16.920 |
| NIR | 0.825 | 11.670 | 0.942 | 33.140 |
| VIS | 0.732 | 3.344 | 0.936 | 38.820 |

**Table 2**. Summary of linear regression for the rocket salad samples in terms of microbial contamination (with and without feature selection) and "time-on-shelf" between the actual measurements and the predictions. Parameters *a, and b* are the slope and bias (offset) of the linear regression (*y=ax+b*) between predicted and measured values, RMSE is the root mean square error of the fit, and R-square is the he coefficient of determination, showing the goodness-of-fit.

| Microbial Contamination | a | b | R-square | RMSE |
|---|---|---|---|---|
| FTIR | 0.741 | 1.978 | 0.468 | 0.359 |
| NIR | 0.335 | 5.397 | 0.239 | 0.272 |
| VIS | 0.544 | 3.699 | 0.474 | 0.261 |
| **Microbial Contamination - No Feature Selection Performed** | | | | |
| FTIR | 0.630 | 2.903 | 0.377 | 0.367 |
| NIR | 0.292 | 5.751 | 0.151 | 0.316 |
| VIS | 0.535 | 3.770 | 0.532 | 0.229 |
| Time-on-shelf | a | b | R-square | RMSE |
| FTIR | 0.730 | -1.480 | 0.565 | 40.690 |
| NIR | 0.801 | 19.8 | 0.675 | 30.620 |
| VIS | 0.909 | 4.523 | 0.799 | 29.250 |

**Table 3**. Summary of linear regression for the baby spinach salad samples in terms of microbial contamination (TVC) (with and without feature selection) and "time-on-shelf" between the actual measurements and the predictions. Parameters *a, and b* are the slope and bias (offset) of the linear regression (*y=ax+b*) between predicted and measured values, RMSE is the root mean square error of the fit, and R-square is the he coefficient of determination, showing the goodness-of-fit.

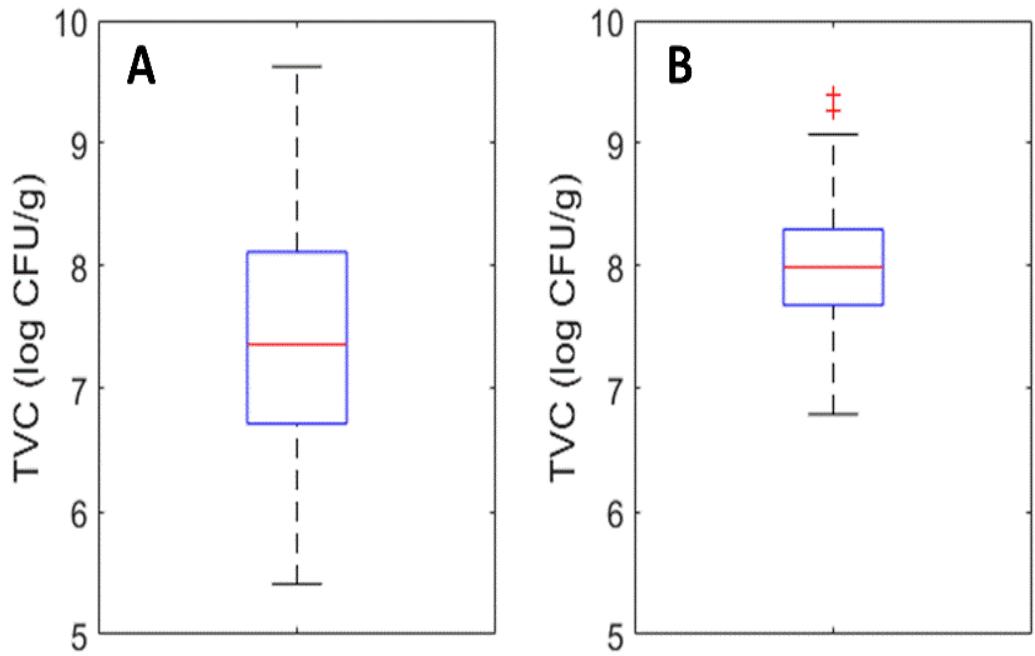

**Figure 1.** Boxplots of the TVC values log CFU/g for A) rocket samples and B) spinach samples.

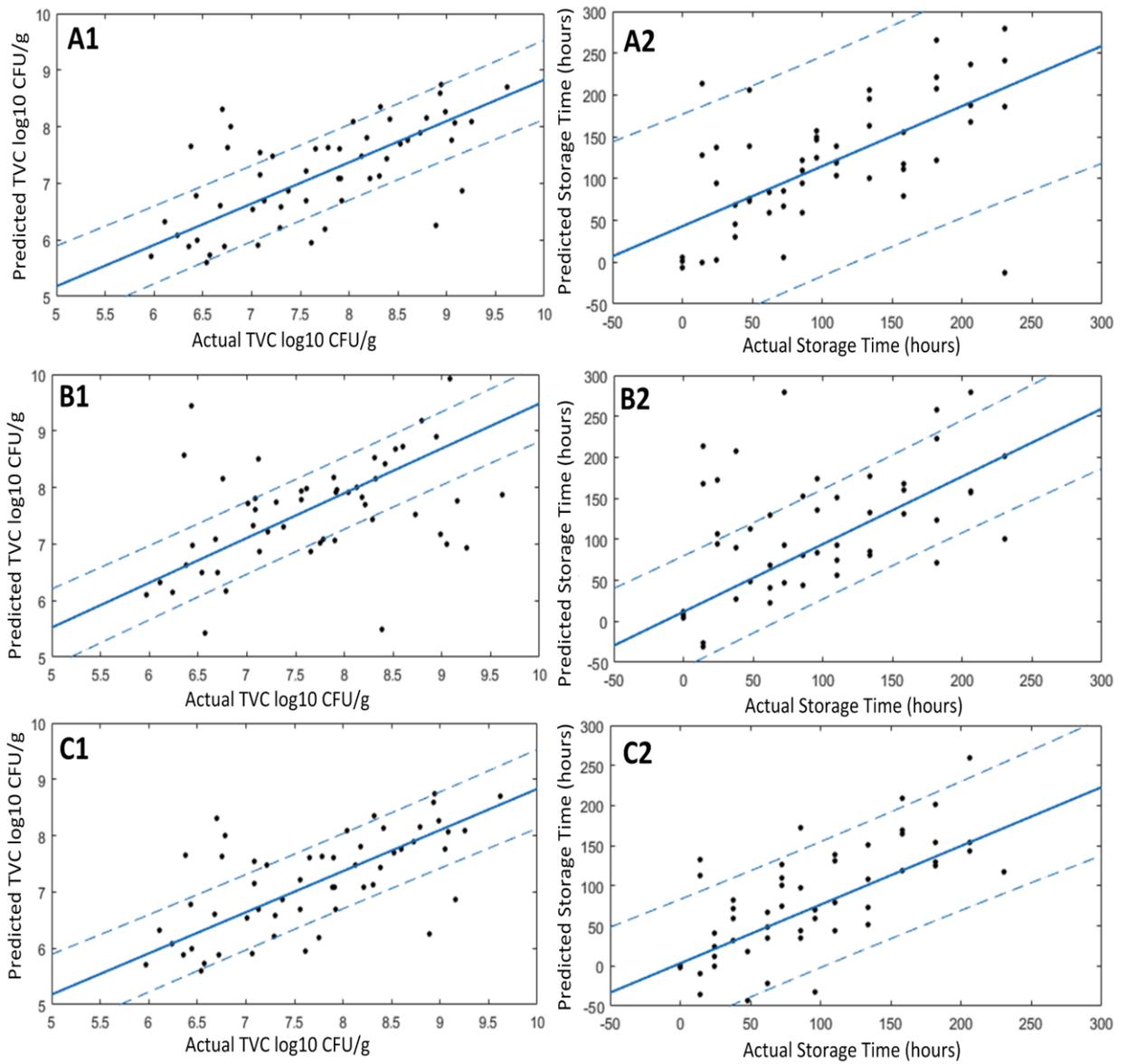

**Figure 2.** Rocket salad samples: A1) FTIR TVC prediction, A2) FTIR "time-on-shelf" prediction, B1) NIR TVC prediction, B2) NIR "time-on-shelf" prediction, C1) VIS TVC prediction, C2) VIS "time-on-shelf" prediction.

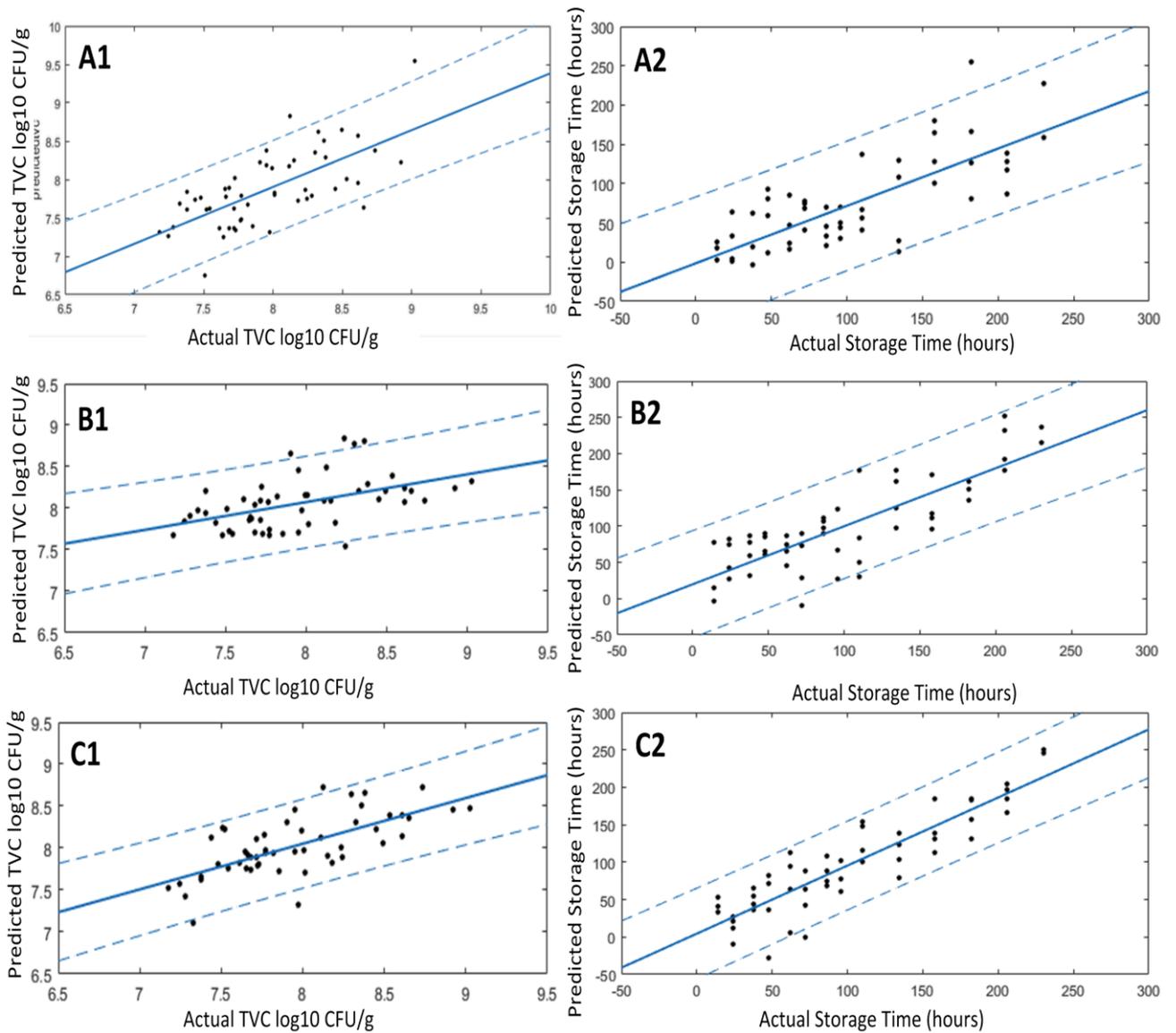

**Figure 3.** Baby spinach salad samples: A1) FTIR TVC prediction, A2) FTIR "time-on-shelf" prediction, B1) NIR TVC prediction, B2) NIR "time-on-shelf" prediction, C1) VIS TVC prediction, C2) VIS "time-on-shelf" prediction.

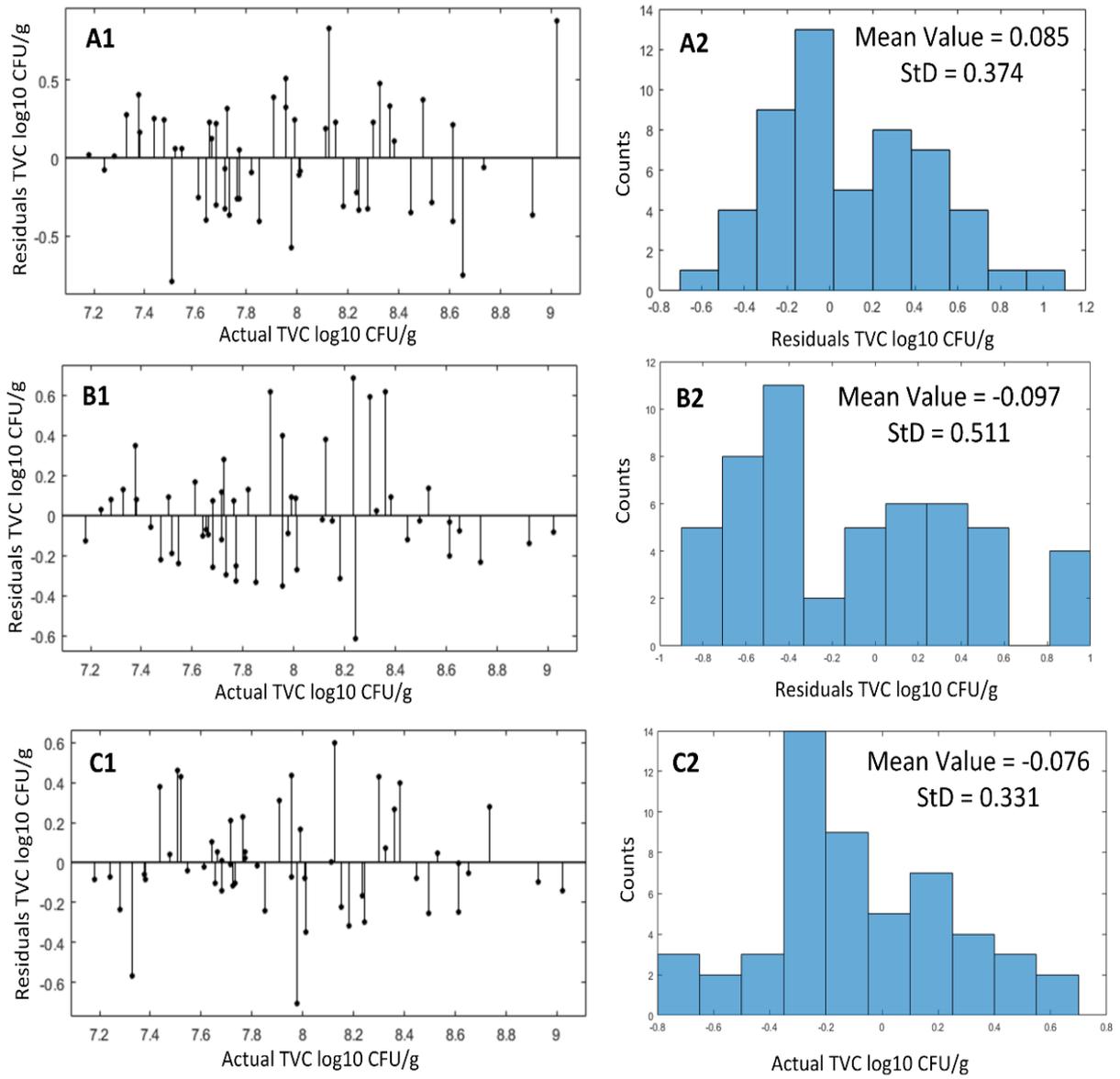

**Figure 4.** Baby spinach salad samples, residual analysis: A1) FTIR residuals, A2) FTIR residuals' histogram, B1) NIR residuals, B2) NIR residuals' histogram, C1) VIS residual, C2) VIS residuals' histogram.